\documentclass[sigconf]{acmart}
\usepackage{libertine}

\AtBeginDocument{%
  }

\copyrightyear{2025}
\acmYear{2025}
\setcopyright{rightsretained}
\acmConference[GECCO '25 Companion]{Genetic and Evolutionary Computation Conference}{July 14--18, 2025}{Malaga, Spain}
\acmBooktitle{Genetic and Evolutionary Computation Conference (GECCO '25 Companion), July 14--18, 2025, Malaga, Spain}
\acmDOI{10.1145/3712255.3726644}
\acmISBN{979-8-4007-1464-1/2025/07}

\usepackage[T1]{fontenc}
\usepackage{graphicx}
\usepackage{dsfont}
\usepackage{bmpsize}
\usepackage{subfigure}
\usepackage{color}
\usepackage{hyperref}
\usepackage{xcolor}
\usepackage{epstopdf}

\usepackage{amsmath}
\usepackage{svg}
\usepackage{newtxtext}
\usepackage{newtxmath}

\usepackage{multirow}
\usepackage{dutchcal}
\usepackage{bbold} 

\usepackage{tikzit}

\tikzstyle{black}=[-, tikzit draw=black, draw=black, fill=none, thick]
\tikzstyle{grey}=[-, tikzit draw={rgb,255: red,128; green,128; blue,128}, draw={rgb,255: red,128; green,128; blue,128}]
\tikzstyle{black dashed}=[-, tikzit draw={rgb,255: red,64; green,64; blue,64}, dashed, draw={rgb,255: red,64; green,64; blue,64}]
\tikzstyle{black dashed capped}=[draw={rgb,255: red,64; green,64; blue,64}, tikzit draw={rgb,255: red,64; green,64; blue,64}, dashed, {|-|}]

\tikzstyle{Box}=[fill=white, draw=black, shape=rectangle, tikzit shape=rectangle, minimum width=2cm, minimum height=1cm]

\tikzstyle{Arrow}=[->]
\tikzstyle{DashedArrow}=[->, dashed]
\tikzstyle{Dashed}=[-, dashed]

\usetikzlibrary{
  positioning 
}

\DeclareMathOperator*{\argmax}{arg\,max}

\renewcommand{\vec}[1]{\mathbf{#1}}

\begin{document}
\title{Efficient Contextual Preferential Bayesian Optimization with Historical Examples}

\author{Farha A. Khan}
\affiliation{%
  \institution{Continental Automotive Components (India) Private Ltd.}
  \country{}
  }
\email{farha.anjum.khan@continental-corporation.com}

\author{Tanmay Chakraborty}
\affiliation{%
  \institution{Continental Automotive Technologies GmbH, Germany}
  \country{}
  }
\email{tanmay.chakraborty@continental-corporation.com}

\author{J{\"o}rg P. Dietrich}
\affiliation{%
  \institution{Continental Automotive Technologies GmbH, Germany}
  \country{}
}
\email{joerg.dietrich@continental-corporation.com}

\author{Christian Wirth}
\affiliation{%
  \institution{Continental Automotive Technologies GmbH, Germany}
  \country{}
}
\email{christian.2.wirth@continental-corporation.com}

\renewcommand{\shortauthors}{Khan et al.}

\begin{abstract}

State-of-the-art multi-objective optimization often assumes a known utility function, learns it interactively, or computes the full Pareto front—each requiring costly expert input.~Real-world problems, however, involve implicit preferences that are hard to formalize. To reduce expert involvement, we propose an offline, interpretable utility learning method that uses expert knowledge, historical examples, and coarse information about the utility space to reduce sample requirements. We model uncertainty via a full Bayesian posterior and propagate it throughout the optimization process. Our method outperforms standard Gaussian processes and BOPE across four domains, showing strong performance even with biased samples, as encountered in the real-world, and limited expert input.
\end{abstract}
\ccsdesc[500]{Mathematics of computing~Bayesian computation}

\keywords{Multi-objective optimization, interpretability, surrogate model}
  


\maketitle  

\section{Introduction}

Preferential Bayesian Optimization (PBO) is a framework for sequential optimization, suited for costly real-world experiments~\citep{PrefBO,
BatchPrefBO}. It models decision-maker preferences through pairwise comparisons of vector-valued outcomes~\citep{lin2022,astudilloqEUBO}, and is widely used in material design, A/B testing, and system tuning.

However, PBO assumes that the decision-maker is always available and optimization is a one-time effort. In real-world, these assumptions often do not hold~\citep{PrefBO}. In this work, we address these limitations by capturing the preferences in advance, thus reducing the need for repeated expert involvement.
We introduce interpretable functional spaces to embed predefined preferences, as proposed by F{\"{u}}rnkranz et al.~\citep{plbook}, allowing decision-makers to validate the optimization.

Our framework is evaluated on benchmarks ZDT3~\citep{Zitzler2000ComparisonOM} and DTLZ2 \citep{DebTLZ}, as well as real-world surrogate CAR~\citep{Carside} and WATER~\citep{WaterRes}.~%
We show the advantage of using a simplified utility function space, informed by prior knowledge, in contrast to two commonly used kernel models without prior knowledge. Furthermore, we use our learned utility functions to guide a Bayesian Optimization (BO) procedure, using EI~\citep{Mockus1975
} and BOPE~\citep{lin2022} exploration strategies. The advantage is demonstrated with unbiased and biased samples.

Our main contributions are: a PBO~\citep{PrefBO} variant with utility uncertainty~\citep{BatchPrefBO} and batch evaluation~\citep{BatchPrefBO} that leverages historical data in a Contextual Bayesian Optimization (CBO) setting; an efficient utility function learning method using prior knowledge and historical samples; an approach to integrate prior knowledge into interpretable function spaces; a demonstration of the importance of uncertainty in misspecified function spaces; and an analysis of the impact of biased samples on utility learning. In contrast to ParEGO~\citep{ParEGO} and other MOBO methods~\citep{MOBO}, our approach learns interpretable utility functions offline, incorporates prior knowledge, and handles uncertainty in misspecified utility spaces.



\section{Problem Setting and Approach}
\label{sec:problem_def}


Our \textbf{problem setting} a variant of the CBO~\citep{CBO} Problem: $f: X \times C \rightarrow \mathds{R}$, with parameter space $X \subset \mathds{R}^{X_{n}}$ and context space $C \subset \mathds{R}^{C_{n}}$. We try to solve $\argmax_{\vec{x}} f(\vec{x})$. 
In contrast to classic CBO, we assume a context-dependent function $g_{c \in C} : X \rightarrow Y$ and a context-independent utility function $e : Y \rightarrow \mathds{R}$. Therefore, $f$ can be defined as $f : e \circ g_c$. Additionally, we assume a dataset $\mathcal{D} \subset Y$. Finally, $e$ is unknown, but we can obtain preferences $\zeta : \mathcal{D} \times \mathcal{D} \in \{\succ,\prec\}$.

This setup reflects common real-world scenarios where we have access to a context-dependent simulator $g_c$ that produces $Y$, qualitatively evaluated through preferences $\{\succ,\prec\}$ rather than numeric scores. The utility function is context independent and we can use historical data $\mathcal{D}$. This leads to three key challenges: (1) defining a suitable space for $e$, (2) estimating $e$ from preferences $\zeta$, and (3) optimizing $\argmax_{\vec{x}} e(g_c(\vec{x}))$. While similar to PBO~\citep{PrefBO
}, our setting differs from CBO since the context $c$ is unobservable.

\subsection{Approach}
Our approach is motivated by several real-world observations: (1) the utility function $e$ should be learned offline to avoid reliance on expert feedback during optimization; (2) historical data $\mathcal{D}$ is typically available; (3) experts can provide prior knowledge about the shape of $e$; and (4) $e$ should capture uncertainty in human feedback. 

We deviate from PBO by estimating $e$ only once
(Sec.~\ref{sec:param_utility}) and incorporate uncertainty by obtaining the full posterior (Sec.~\ref{sec:preference_learning}). We then optimize with the context-specific simulator $g_c$ by solving $\argmax_{\vec{x}} \bar{e}(g_c(\vec{x}), \vec{p}_\zeta)$ with posterior samples $\vec{p}_\zeta \sim \Pr(\vec{p}|\zeta)$ (Sec.~\ref{sec:optimization}). Additionally, we encode expert priors within an interpretable, parametric function space $e$.

\begin{table*}[htb]
\def\arraystretch{1.2}
{\setlength{\tabcolsep}{0.5em}
\setlength{\abovecaptionskip}{5pt}
\centering

\begin{tabular}{|l|c|c|c|c|c|c|c|c|c|c|c|c|}
\cline{1-13} 
&  \multicolumn{3}{c|}{ZDT3}   & \multicolumn{3}{c|}{DTLZ2}   &  \multicolumn{3}{c|}{CAR}   & \multicolumn{3}{c|}{WATER}  \\ 
\cline{1-13} 
                                                 
$N$ & 10 & 20 & 50 & 10 & 20 & 50 & 10 & 20 & 50 & 10 & 20 & 50  \\ \hline \hline

Lin  & 0.54$_\text{-.04}^\text{+.04}$  & 0.56$_\text{-.05}^\text{+.06}$   & 0.64$_\text{-.01}^\text{+.00}$  & 0.71$_\text{-.03}^\text{+.03}$   & 0.70$_\text{-.01}^\text{+.02}$  & 0.81$_\text{-.06}^\text{+.04}$  & 0.18$_\text{-.16}^\text{+.19}$   & 0.30$_\text{-.19}^\text{+.13}$  & 0.46$_\text{-.06}^\text{+.04}$   & 0.50$_\text{-.28}^\text{+.18}$  & 0.68$_\text{-.09}^\text{+.05}$  & 0.75$_\text{-.03}^\text{+.02}$ \\ \hline

Adap & 0.48$_\text{-.09}^\text{+.17}$  & 0.83$_\text{-.19}^\text{+.12}$  & 0.95$_\text{-.05}^\text{+.02}$  & 0.75$_\text{-.05}^\text{+.08}$  & 0.76$_\text{-.05}^\text{+.09}$  & 0.91$_\text{-.04}^\text{+.03}$   & 0.43$_\text{-.26}^\text{+.26}$   & 0.80$_\text{-.06}^\text{+.06}$  & \textbf{0.86}$_\textbf{-.04}^\textbf{+.03}$  & 0.56$_\text{-.21}^\text{+.11}$   & 0.64$_\text{-.1}^\text{+.11}$  & 0.78$_\text{-.04}^\text{+.02}$   \\ \hline

Inf & \textbf{0.88}$_\textbf{-.02}^\textbf{+.03}$  & \textbf{0.93}$_\textbf{-.03}^\textbf{+.02}$    & \textbf{0.94}$_\textbf{-.00}^\textbf{+.00}$   & 0.82$_\text{-.11}^\text{+.1}$  & \textbf{0.92}$_\textbf{-.05}^\textbf{+.03}$  & \textbf{0.95}$_\textbf{-.01}^\textbf{+.01}$ & \textbf{0.63}$_\textbf{-.15}^\textbf{+.13}$  & \textbf{0.81}$_\textbf{-.06}^\textbf{+.05}$  & \textbf{0.86}$_\textbf{-.05}^\textbf{+.02}$  & 0.64$_\text{-.06}^\text{+.07}$  & 0.68$_\text{-.05}^\text{+.09}$  & 0.79$_\text{-.06}^\text{+.04}$  \\ \hline

 SVM      & 0.82$_\text{-.13}^\text{+.06}$  & 0.86$_\text{-.08}^\text{+.05}$  & 0.86$_\text{-.04}^\text{+.03}$  & \textbf{0.86}$_\textbf{-.05}^\textbf{+.04}$  & 0.89$_\text{-.06}^\text{+.02}$   & 0.91$_\text{-.01}^\text{+.00}$ & 0.48$_\text{-.17}^\text{+.12}$  & 0.62$_\text{-.07}^\text{+.09}$  & 0.72$_\text{-.06}^\text{+.03}$   & \textbf{0.74}$_\textbf{-.13}^\textbf{+.1}$  & \textbf{0.84}$_\textbf{-.04}^\textbf{+.04}$  & \textbf{0.92}$_\textbf{-.02}^\textbf{+.01}$  \\ 

GP     & 0.71$_\text{-.15}^\text{+.15}$  & 0.74$_\text{-.13}^\text{+.12}$   & 0.75$_\text{-.1}^\text{+.09}$  & 0.81$_\text{-.13}^\text{+.06}$  & 0.84$_\text{-.08}^\text{+.05}$  & 0.91$_\text{-.02}^\text{+.03}$  & 0.49$_\text{-.21}^\text{+.17}$  & 0.64$_\text{-.04}^\text{+.05}$  & 0.72$_\text{-.06}^\text{+.05}$  & 0.69$_\text{-.19}^\text{+.15}$  & 0.73$_\text{-.15}^\text{+.11}$   & 0.90$_\text{-.05}^\text{+.04}$ \\  \hline

\end{tabular}
\caption{Kendall's Tau ranking similarity between $e^*$ and surrogate $e$ (higher is better), with error limits indicated by - and +.
}
\label{tbl:kendall_rn}}
\end{table*}

\begin{table*}[htb]
\def\arraystretch{1.2}
{\setlength{\tabcolsep}{0.5em}
\setlength{\abovecaptionskip}{5pt}
\centering

\begin{tabular}{|l|c|c|c|c|c|c|c|c|c||c|c|c|}
\cline{1-13} 
&  \multicolumn{3}{c|}{ZDT3}   & \multicolumn{3}{c|}{DTLZ2} &  \multicolumn{3}{c|}{CAR}   & \multicolumn{3}{c|}{WATER}  \\ 
\cline{1-13} 
                                                 
$N$ & 10 & 20 & 50 & 10 & 20 & 50 & 10 & 20 & 50 & 10 & 20 & 50 \\ \hline \hline

Lin & 0.54$_\text{-.03}^\text{+.02}$   & 0.50$_\text{-.01}^\text{+.02}$  & 0.54$_\text{-.04}^\text{+.03}$  & 0.71$_\text{-.01}^\text{+.03}$  & 0.72$_\text{-.05}^\text{+.07}$  & 0.79$_\text{-.06}^\text{+.05}$ & 0.12$_\text{-.18}^\text{+.11}$   & 0.13$_\text{-.13}^\text{+.22}$   & 0.03$_\text{-.07}^\text{+.09}$  & 0.43$_\text{-.07}^\text{+.04}$    & 0.50$_\text{-.32}^\text{+.1}$   & 0.30$_\text{-.06}^\text{+.16}$ \\ \hline

Adap & 0.44$_\text{-.02}^\text{+.02}$   & 0.40$_\text{-.02}^\text{+.05}$   & 0.47$_\text{-.05}^\text{+.3}$   & 0.80$_\text{-.04}^\text{+.08}$  & 0.78$_\text{-.18}^\text{+.11}$  & \textbf{0.81}$_\textbf{-.05}^\textbf{+.04}$ & -.01$_\text{-.26}^\text{+.12}$  & -.07$_\text{-.08}^\text{+.14}$   & -.07$_\text{-.1}^\text{+.19}$   & 0.33$_\text{-.07}^\text{+.1}$  & 0.46$_\text{-.16}^\text{+.11}$  & 0.45$_\text{-.15}^\text{+.17}$ \\ \hline

Inf & \textbf{0.88}$_\textbf{-.00}^\textbf{+.01}$   & \textbf{0.87}$_\textbf{-.00}^\textbf{+.01}$  & \textbf{0.88}$_\textbf{-.01}^\textbf{+.01}$  & \textbf{0.87}$_\textbf{-.02}^\textbf{+.03}$  & \textbf{0.77}$_\textbf{-.08}^\textbf{+.09}$  & 0.79$_\text{-.04}^\text{+.07}$  & \textbf{0.19}$_\textbf{-.13}^\textbf{+.11}$   & 0.02$_\text{-.03}^\text{+.06}$    & 0.06$_\text{-.1}^\text{+.25}$  & 0.47$_\text{-.02}^\text{+.02}$  & \textbf{0.58}$_\textbf{-.07}^\textbf{+.04}$  & \textbf{0.61}$_\textbf{-.02}^\textbf{+.03}$  \\ \hline

SVM      & 0.61$_\text{-.01}^\text{+.01}$  & 0.62$_\text{-.01}^\text{+.01}$   & 0.62$_\text{-.00}^\text{+.01}$  & 0.49$_\text{-.12}^\text{+.18}$  & 0.55$_\text{-.06}^\text{+.06}$  & 0.71$_\text{-.08}^\text{+.07}$  & 0.11$_\text{-.16}^\text{+.1}$   & 0.12$_\text{-.04}^\text{+.06}$   & 0.13$_\text{-.03}^\text{+.04}$   & \textbf{0.50}$_\textbf{-.04}^\textbf{+.02}$  & 0.49$_\text{-.09}^\text{+.04}$   & 0.49$_\text{-.11}^\text{+.1}$  \\ 

GP    & 0.61$_\text{-.01}^\text{+.02}$    & 0.61$_\text{-.00}^\text{+.00}$   & 0.61$_\text{-.00}^\text{+.01}$  & -.07$_\text{-.39}^\text{+.4}$  & 0.24$_\text{-.27}^\text{+.25}$  & 0.59$_\text{-.28}^\text{+.17}$  & 0.15$_\text{-.15}^\text{+.08}$   & \textbf{0.18}$_\textbf{-.06}^\textbf{+.07}$   & \textbf{0.16}$_\textbf{-.08}^\textbf{+.04}$  & 0.35$_\text{-.28}^\text{+.35}$  & 0.44$_\text{-.56}^\text{+.32}$  & 0.41$_\text{-.25}^\text{+.22}$   \\  \hline

\end{tabular}
\caption{Kendall's Tau ranking similarity between $e^*$ and surrogate $e$ (higher is better), based on biased samples.
}
\label{tbl:kendall_biased}}
\end{table*}

\begin{figure}[ht]
\centering



\begin{tikzpicture}[scale=1.5,left]
	\begin{pgfonlayer}{nodelayer}
		\node [style=none] (0) at (0, 0) {};
		\node [style=none] (1) at (3, 0) {};
		\node [style=none] (2) at (0, 3) {};
		\node [style=none] (3) at (0, 0.5) {};
		\node [style=none] (4) at (0.5, 0.5) {};
		\node [style=none] (5) at (2.5, 2) {};
		\node [style=none] (6) at (3, 2) {};
		\node [style=none] (7) at (2.5, 0.5) {};
		\node [style=none] (9) at (2.75, 1) {a};
		\node [style=none] (12) at (3.5, 0) {$y$};
		\node [style=none] (13) at (0, 3.25) {$e(y)$};
		\node [style=none] (14) at (-0.75, 0.125) {b};
		\node [style=none] (17) at (0, 0.125) {};
		\node [style=none] (18) at (0, 0) {};
		\node [style=none] (19) at (2.5, 2) {};
		\node [style=none] (20) at (1.25, 1.75) {};
		\node [style=none] (21) at (1.25, 1.75) {$\alpha$};
		\node [style=none] (22) at (2.5, 2) {};
		\node [style=none] (23) at (2.5, 2) {};
		\node [style=none] (24) at (2.5, 2) {};
		\node [style=none] (25) at (1.25, 1.5) {};
		\node [style=none] (26) at (1.5, 1.25) {};
		\node [style=none] (27) at (-0.35, 0.5) {};
		\node [style=none] (28) at (-0.375, 0) {};
            %
            %
            \node [style=none] (32) at (-0.05, -0.05) {};
		\node [style=none] (33) at (0.1, 0.1) {};
            \node [style=none] (34) at (3., -0.1) {};
		\node [style=none] (35) at (3., 0.1) {};
            %
            %
	\end{pgfonlayer}
	\begin{pgfonlayer}{edgelayer}
		\draw [style=grey] (2.center) to (0.center);
		\draw [style=grey] (0.center) to (1.center);
		\draw [style=black] (3.center) to (4.center);
		\draw [style=black, bend left] (4.center) to (5.center);
		\draw [style=black] (5.center) to (6.center);
		\draw [style=black dashed] (4.center) to (17.center);
		\draw [style=black dashed] (4.center) to (19.center);
		\draw [style=black dashed capped] (19.center) to (7.center);
		\draw [style=black dashed capped] (25.center) to (26.center);
		\draw [style=black dashed capped] (27.center) to (28.center);
            %
            %
            %
            %
	\end{pgfonlayer}
\end{tikzpicture}
\begin{tikzpicture}[scale=1.5,right]
	\begin{pgfonlayer}{nodelayer}
		\node [style=none] (0) at (0, 0) {};
		\node [style=none] (1) at (3, 0) {};
		\node [style=none] (2) at (0, 3) {};
		\node [style=none] (3) at (0, 0.5) {};
		\node [style=none] (4) at (0.5, 0.5) {};
		\node [style=none] (5) at (2.5, 2) {};
		\node [style=none] (6) at (3, 2) {};
            \node [style=none] (7) at (-0.075, 2.) {};
		\node [style=none] (8) at (0.075, 2.) {};
		\node [style=none] (9) at (1.5, -0.25) {};
		\node [style=none] (10) at (1.5, 0.5) {d};
		\node [style=none] (12) at (3.5, 0) {$y$};
		\node [style=none] (13) at (0, 3.25) {$e(y)$};
            \node [style=none] (14) at (0, -0.25) {$y_\text{min}$};
            \node [style=none] (15) at (3., -0.25) {$y_\text{max}$};
		\node [style=none] (16) at (-0.25, 0.5) {0};
		\node [style=none] (17) at (-0.25, 2) {1};
		\node [style=none] (18) at (0.5, 0.25) {};
		\node [style=none] (19) at (2.5, 0.25) {};
		\node [style=none] (20) at (0.5, 1) {};
		\node [style=none] (21) at (0, 1) {};
		\node [style=none] (22) at (0.25, 1.25) {b};
		\node [style=none] (23) at (1.25, 1.75) {$\alpha$};
		\node [style=none] (24) at (1.25, 1.5) {};
		\node [style=none] (25) at (1.5, 1.25) {};
            \node [style=none] (26) at (-0.05, -0.05) {};
		\node [style=none] (27) at (0.1, 0.1) {};
            \node [style=none] (28) at (3., -0.1) {};
		\node [style=none] (29) at (3., 0.1) {};
	\end{pgfonlayer}
	\begin{pgfonlayer}{edgelayer}
		\draw [style=grey] (2.center) to (0.center);
		\draw [style=grey] (0.center) to (1.center);
		\draw [style=black] (3.center) to (4.center);
		\draw [style=black, bend left] (4.center) to (5.center);
		\draw [style=black] (5.center) to (6.center);
            \draw [style=grey] (7.center) to (8.center);
		\draw [style=black dashed capped, in=180, out=0] (21.center) to (20.center);
		\draw [style=black dashed capped] (18.center) to (19.center);
		\draw [style=black dashed capped] (4.center) to (5.center);
		\draw [style=black dashed capped] (24.center) to (25.center);
            \draw [style=grey] (26.center) to (27.center);
            \draw [style=grey] (28.center) to (29.center);
	\end{pgfonlayer}
\end{tikzpicture}
  



\caption{Common ({\bf Left}) and our ({\bf Right}) monotone function.
}
\label{fig:fn_both}
\end{figure}
\subsubsection{Parametric Utility Functions}
\label{sec:param_utility}
Most approaches focus on utility functions $e$ with a high expressiveness, such as Gaussian Processes~\citep{PrefGP}, that are not interpretable and prior knowledge cannot be incorporated.

We capture each output dimension interdependently with a possibly non-linear utility function $e_i(y_i)$ and perform a linear aggregation.~%
This allows the practitioner to define independent priors, as in~\citep{astudilloExo}, without requiring repeated expert involvement. Additionally, our space is deemed interpretable, due to the linear aggregation and bounding it to $[0,1]$.


In the following, we omit the subscript $i$ for readability. A common class of $[0,1]$-bounded monotonic functions is defined as (Fig.~\ref{fig:fn_both}):
\begin{equation}
\label{eq:common_monotone}
e(y) = \min(1,\max(0,a \cdot y^{\alpha}+b)),
\end{equation}
where $a$ is the gradient, $b$ the bias, and $\alpha$ the exponent. This formulation has interdependent parameters, making it difficult to impose meaningful constraints: (1) $\max e(y)$ may be $<1$ and $\min e(y) > 0$; (2) priors for $a$ and $b$ depend on the scale of $y$; (3) $a$ controls both monotonicity and steepness, preventing independent priors; and (4) the points where $e(y)=0$ or $e(y)=1$ are only implicitly defined.

Our monotonic utility function (Fig.~\ref{fig:fn_both} right) is defined as:
\begin{equation}
\label{eq:utilityPowerFn}
\begin{aligned}
u_s(y) &= \frac{y - y_\mathrm{min}}{y_\mathrm{max} - y_\mathrm{min}}, \\
e(y) &= \min\left(1, \max\left(0, m \cdot \left(\frac{u_s(y) - b}{d}\right)^{\alpha} \right)\right),
\end{aligned}
\end{equation}
with parameters: $y_\mathrm{min}$ and $y_\mathrm{max}$ defining bounds where $e(y) = 0$ or $1$; $b \in [0, 1]$ as the start of the monotonic segment; $d \in [0, 1]$ its length; $\alpha \in [p_{\min}, p_{\max}]$ controlling non-linearity; and $m \in \{-1, 1\}$ indicating direction. This formulation ensures $[0,1]$ bounds. 
$p_{\min}$ and $p_{\max}$ are the curvature limits and vary depending on the shape of $e$.
%
The full utility function $e(\vec{p}, \vec{y}) = \sum_{y_i \in \vec{y}} w_i\, e_i(y_i, \vec{p}_i),$ where $\vec{p}$ stacks parameters $\vec{p}_i$ for each $e_i$, and weights $w_i \in \mathds{R}^+$.~
%
To ensure stability, we normalize the weights with $\|\vec{w}\|_1 = 1$ after learning. 
We recommend fixing $y_\mathrm{min}$ and $y_\mathrm{max}$ using expert knowledge or observed values, as optimizing them is computationally challenging and bias prone.~
Since we make no assumptions about the form of $e_i$, alternative, even non-linear functions, are possible. 

\subsubsection{Preference Learning}
\label{sec:preference_learning}
Due to the interpretability of $\vec{p}$, it is possible to partially fix the parameters, based on domain knowledge. We learn the remaining parameters of $e$ via preference learning, E.g. $e(y_0) > e(y_1) \rightarrow y_0 \succ y_1$.
We use a probabilistic fulfillment function $\Pr(\vec{p} \vert r_i) = \mathrm{sigmoid}(e(\vec{y}_0, \vec{p}) - e(\vec{y}_1, \vec{p}))$ with $r_i \in \zeta$~\citep{SigPreference}.
Hence, we can find the optimal utility function parameters $\arg\max_{\vec{p}} \Pr(\vec{p} \vert \zeta)$ by
\begin{equation}
\begin{aligned}
\label{eq:scoreOpt}
\vec{p}_{\zeta} &= \arg\max_{\vec{p}} \sum^{\vert\zeta\vert}_{i} \log(\mathrm{sigmoid}(e(\vec{y}_{i_0}, \vec{p}) - e(\vec{y}_{i_1}, \vec{p}))
\end{aligned}
\end{equation}
with $\vec{y}_{i_0}$ and $\vec{y}_{i_1}$ as the preferred (and dominated) examples of preference $r_i \in \zeta$.
However, we use a full Bayesian posterior distribution over utility functions to prevent overfitting with limited data and ensure robustness against errors.


We approximate the posterior distribution $\Pr(e(\vec{y}) \vert \zeta)$ by obtaining function parameter posterior samples 
\begin{equation}
\label{eq:scoreOptDist}
\mathcal{p}_{\zeta} \sim \prod^{\vert\zeta\vert}_{i} \mathrm{sigmoid}(e(\vec{y}_{i_0}, \vec{p}) - e(\vec{y}_{i_1}, \vec{p}))
\end{equation}
via No-U-Turn sampling~\citep{NUTS}. $\mathcal{p}_\mathrm{opt}$ is then a set of samples of $\vec{p} \in P$, empirically approximating the full posterior distribution,.

\subsubsection{Optimization}
\label{sec:optimization}
As we assume samples $\mathcal{p}_\mathrm{opt}$, one can use the parameter mean $\bar{\mathcal{p}}_{\zeta}$ and
$\vec{x}^* = \argmax_{\vec{x}} e(f(\vec{x}), \bar{\mathcal{p}}_{\zeta})$
However, this point estimate may result in overfitting (cf.~\ref{sec:preference_learning}). 
Therefore, we optimize $\vec{x}^{*} = \argmax_{\vec{x}} \bar{e}(f(\vec{x}), \mathcal{p}_{\zeta})$ with $\bar{e}(f(\vec{x}), \mathcal{p}_{\zeta})$ as the mean of the utility. This turns our optimization target into the full Bayesian mean estimate $\Pr(\vec{x} \vert  \zeta)$. For which we can find the solution with conventional BO methods~\citep{BOsurvey}.

\section{Evaluation}
\label{sec:evaluation}
We evaluate the benchmark problems ZDT3~\citep{Zitzler2000ComparisonOM}, DTLZ2~\citep{DebTLZ} and the constrained, real-world surrogates CAR~\citep{Carside} and WATER~\citep{WaterRes}. 

We use an unknown function $e^*$ as expert surrogate (which may not be representable by $e(\vec{y},\vec{p})$, (cf. Sec~\ref{sec:param_utility}). We split the domain of $X$ into a $80\%$ train and $20\%$ and define the context by fixing $x_0$ to $N$ evenly distributed values. 
The remaining $x_i \in \vec{x}$ are sampled randomly.
We obtain $N \cdot (N-1) / 2$ preferences by $\zeta = \forall \vec{y}_{i}, \vec{y}_{j} \in \mathcal{D} \times\mathcal{D}: \mathds{1}[e^*(\vec{y}_{0}) > e^*(\vec{y}_{1})]$, excluding symmetry.

The defined expert functions for four domains are: $e^*$(ZDT3)=$2 \cdot \min(0.5,f_1) \cdot 2 \cdot \min(0.5,f_2)$, $e^*$(DTLZ2)=$0.3 \cdot f_1 \cdot f_2+ 0.7 \cdot (2\cdot\min(0.5,f_3))^2$, $e^*$(CAR)=$0.3 \cdot f_1 \cdot f_2 + 0.7 \cdot (2 \cdot \min(0.5,f_3))^2$ and $e^*$(WATER)=$0.2 \cdot f_1 \cdot f_2+ 0.4 \cdot f_3 \cdot f_4+ 0.4 \cdot (2 \cdot (\min(0.5,f_5))^2$. These functions include dependencies between dimensions, non-linearities and truncated terms, therefore, mimick complex, expert evaluations.

The trade-off factors are arbitrarily chosen, but pre-evaluated to not result in trivial, optimal solutions. Note that $e$ is misspecified wrt. to the multiplicative and quadratic terms of $e^*$.
%

%
\paragraph{\textbf{Surrogate Utility Function}}
\label{sec:exp_utility}
We differentiate four surrogate utility functions:
\textit{Linear} assumes a linear function $e(\vec{y}) = \sum_{y_i \in \vec{y}} w_i  y_i$.

\textit{Informed} use the utility function space as defined in Sec.~\ref{sec:param_utility} with $y_\mathrm{min}$, $y_\mathrm{max}$ and $m$ informed by $e^*$ (~Sec.~\ref{sec:evaluation}). 

\textit{Adaptable} uses the same function space, but $m$ is learned, ($y_\mathrm{min}$, $y_\mathrm{max}$) set to the general bounds (e.g. for ZDT3 $f_2$: $y_\mathrm{min} = 0$, $y_\mathrm{max} = 8.0$) and no normalization (Sec.~\ref{sec:param_utility}).~Therefore, \emph{Adaptable} can learn the same utility function as \emph{Informed}, but does not have access to the explicit expert information ($m, y_\mathrm{min}, y_\mathrm{max}$) and is not interpretable. 

\textit{Kernel} settings employ an \emph{RBF Kernel}~\citep{rbfbook
} in a Support Vector Machine~\citep[SVM;][]{SVMrank} or Gaussian Process~\citep[GP;][]{PrefGP} model~\citep{PrefBO,BatchPrefBO}.

We estimate our utility function using the full posterior Eq.~\ref{eq:scoreOptDist}.

\paragraph{\textbf{Experiment Setting}}

We show that our system is able to recover the (unknown) expert function $e^*$. We run 4 parallel sampling chains (Sec.~\ref{sec:preference_learning}) with default parameters\footnote{\url{https://www.tensorflow.org/probability/api_docs/python/tfp/mcmc/NoUTurnSampler}} and step size $0.0005$ till $\textit{PSR}<1.1$ or 1M samples are generated. The \emph{Posterior} estimate is a 20k random subset.
%
The GP\footnote{\url{https://github.com/chariff/GPro}} and SVM\footnote{\url{https://www.cs.cornell.edu/people/tj/svm_light/svm_rank.html}} use default hyperparameters. The approximation $e$ is evaluated via Kendall's $\tau$, w.r.t. $e^*$. All results are averaged over 10 trials.



\paragraph{\textbf{Black-Box Multi-Objective Optimization}}
\label{sec:exp_mobo}
%
We also use our learned utility functions as optimization target for BO~\citep{BOsurvey}. We use the hyperparameters from the BOPE~\citep{lin2022} implementation\footnote{\url{https://botorch.org/tutorials/bope}}.~
%
Expected Improvement (EI)~\cite{Mockus1975} and qNEIUU~\citep{lin2022} are used as acquisition function.
Note, in contrast to BOPE, we do not collect additional preferences during the optimization loop.
A GP as function space was also proposed by \citep{PrefBO,BatchPrefBO}, making it a common baseline. The BO method was run for 800 iterations.

\begin{figure*}[htb]
\centering
%
\includegraphics[width=\textwidth]{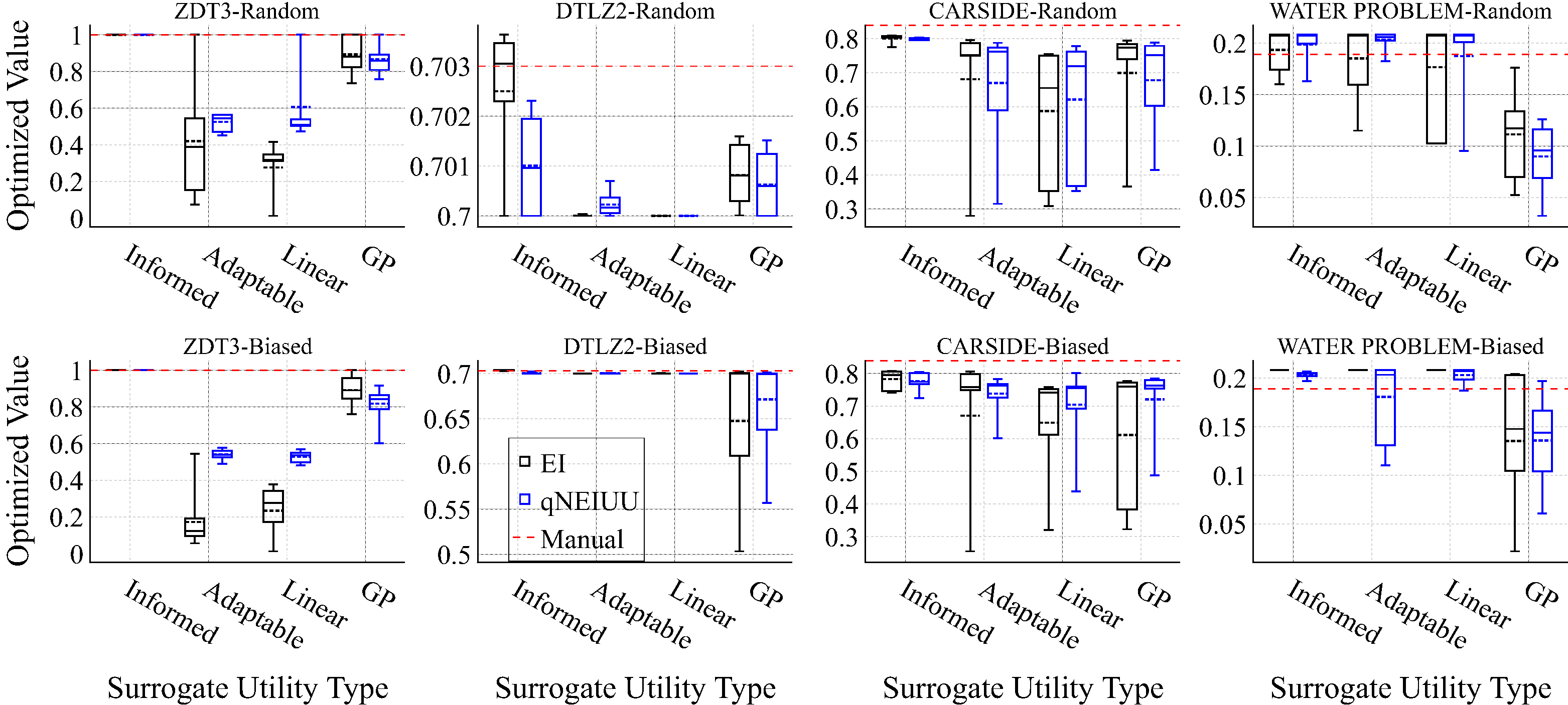}
\caption{Mean (dashed) and median (solid) $e^*$ score of the best results learned from random ({\bf top}) and biased samples ({\bf bottom}).
}
\label{fig:res1}
\end{figure*}

\paragraph{\textbf{Results: Surrogate Utility Function}}
\label{sec:result_utility}

Table~\ref{tbl:kendall_rn} shows that our \emph{Informed} method outperforms others and mostly requires fewer samples, thus capturing prior knowledge more efficiently.~
Utility functions with parameter dependencies (\emph{Adaptable} and \emph{Informed}) perform similarly, but \emph{Adaptable} struggles with limited data.~
%
%
The \emph{Linear} baseline is too simple to approximate the true utility function.~


\paragraph{\textbf{Ablation Study Results: Biased Samples}}
\label{sec:results_biased_utility}

Real-world, historical results $\mathcal{D}$ are usually biased because only optimized outcomes have been stored. 
To simulate this, we create $N$ sets with 1k random examples and choose $\argmax e^*(\vec{y})$. All results are 10 trial averages.~

Table~\ref{tbl:kendall_biased} shows that learning from biased samples reduces performance because they only cover parts of the space of $\vec{y}$.
%
%
Increasing the sample count can even strengthen that effect.
%
However, the \emph{Informed} method is still superior,  except for \emph{CAR}.

Independent of the biasing, incorporating prior knowledge still leads to improved utility functions, especially in the small sample ($\sim$10) regime. Furthermore, our \emph{Informed} method can achieve the same or better results compared to GP. Without prior knowledge, simpler function spaces cannot compete, but all advancements together allow us to compute high-quality solutions with an interpretable and informed optimization target.
%
%
%
\paragraph{\textbf{Results: Black-Box Multi-Objective Optimization}}
\label{sec:result_mobo}

Fig.~\ref{fig:res1} shows the results in terms of $e^*$ (Sec.~\ref{sec:exp_mobo}). The trials use the utility functions resulting from Sec.~\ref{sec:result_utility} (\emph{Random} and \emph{Biased}), with $N=10$. \emph{Manual} is an approximate upper bound, obtained by selecting $\max e^*$ from 10M random samples. 

The results show that our approach (\emph{Informed}) outperforms GP (with BOPE~\citep{lin2022} and qNEIUU strategy). Furthermore, it indicates that our methods perform comparably, independent of the acquisition function (EI and qNEIUU). Additionally, \emph{Informed} dominates the results in all domains, which shows that incorporating prior knowledge is helpful.
In some cases, the results outperform the upper bound, indicating that the problems are too complex for simple optimization procedures. Therefore, we can see that our method is not restricted to simple problem domains.
Interestingly, biased samples are not an issue in the \emph{Informed} setting, likely because it is sufficient to guide the optimizer into an area with high-utility solutions.
%

\section{Conclusion}
\label{sec:conclusion}
We propose a novel method for learning interpretable utility functions from preferences, prior knowledge, and historical data with minimal expert input.~Our approach outperforms standard GP models in several domains and supports biased or limited data. By modeling uncertainty and incorporating expert priors, we address challenges of misspecified function spaces. Future work includes exploring more flexible yet interpretable function spaces and applying the method to complex real-world problems.

\begin{acks}
This study was supported by BMBF Project hKI-Chemie: humancentric AI for the chemical industry, FKZ 01|S21023D, FKZ 01|S21023G and Continental AG.
\end{acks}

\bibliographystyle{ACM-Reference-Format}
\bibliography{bibliogaphy}


\end{document}